\documentclass[conference,a4paper]{IEEEtran}
\IEEEoverridecommandlockouts

\usepackage[hidelinks]{hyperref}
\usepackage[cmex10]{amsmath}%American Math Society(AMS) math formatting
\usepackage{amssymb,amsfonts}%AMS extra symbols and fonts
\usepackage{dblfloatfix}%fix double column figure ordering and placement
\usepackage{placeins}
\usepackage[ruled,vlined]{algorithm2e}
\usepackage{graphicx}
\graphicspath{{Figures/PDF/}{Figures/PNG/}}
\usepackage{float}
\usepackage{booktabs}
\usepackage{siunitx}
\usepackage[numbers,compress]{natbib}
\usepackage{texnames}
\usepackage{bm,bbm}
\usepackage{orcidlink}
\usepackage{mathtools}
\usepackage{subcaption}
\begin{document}

\title{\uppercase{Spectral Super-Resolution using Spatial-Spectral Residual Operator Networks}
}

\author{	\IEEEauthorblockN{Seokhyun Chin
	\IEEEauthorblockA{\textit{California Institute of Technology}\\
		1200 E California Boulevard, Pasadena, CA, USA\\
		schin@caltech.edu}
}}

\maketitle
\begin{abstract}
	Spectral super-resolution of multispectral satellite images can enable high temporal- and spatial-resolution hyperspectral satellite imagery at a modest cost, significantly increasing the applicability of hyperspectral remote sensing. This task is inherently ill-posed, making it well-suited for deep learning-based methods. In this study, the spectral super-resolution task is framed as an operator learning problem, and SSRON is proposed as a Deep Operator Network that effectively learns function-to-function mappings from downsampled spectra to continuous spectra. The model is trained to super-resolve Sentinel-2A-like multispectral imagery to EMIT images. Compared to baseline models, SSRON achieves superior performance across all metrics. The model also demonstrates zero-shot spectral super-resolution capability by predicting bands unseen during training. Furthermore, its continuous-output formulation suggests the potential to estimate spectra at finer wavelength intervals than the native sensor. These results suggest the potential of SSRON and establishes operator learning as a promising direction for spectral super-resolution. 
\end{abstract}


\section{Introduction}
Hyperspectral satellite images (HSIs) have become important tools in earth observation. By capturing hundreds of spectral bands, they provide detailed spectral information that is unavailable in conventional satellite imagery. As a result, HSIs have been widely used in applications such as weather forecasting \cite{Smith2020}, precision agriculture \cite{Singh2020}, and water quality monitoring \cite{Chander2020}. However, their broader use is limited by low spatio-temporal resolution and high data acquisition costs \cite{Imran2024}.

To address these limitations, three main approaches have been developed. Spatial Super Resolution seeks to directly enhance the spatial resolution of an HSI; HSI sharpening fuses HSIs with high-resolution multispectral satellite images (MSIs) to improve spatial detail; Spectral Super Resolution (SSR) seeks to enhance the spectral resolution of an MSI to obtain an HSI. HSI sharpening suffers from the lack of high-spatial-resolution HSI reference data, and Spatial Super Resolution cannot overcome the low temporal resolution of HSI or the high cost of HSI acquisition \cite{Xie2024}. Consequently, SSR provides a more practical and flexible alternative.

However, SSR is an ill-posed inverse problem in which hundreds of HSI bands are recovered from only a limited number of MSI bands. Traditional SSR is based on shallow encoding methods such as dictionary-based methods
\cite{Arad2016, Han2019, Fotiadou2019}. With the emergence of deep learning methods, research has increasingly shifted toward their use. Specifically, variations of convolutional neural networks, generative adversarial networks, and transformer-based architectures have shown strong performance \cite{Li2022, Mu2022, SSRAN, Du2023, Gonzalez2025, Vandal2022}. 

Neural operators have emerged as a deep learning framework that learns mappings between infinite-dimensional function spaces \cite{Lu2021, Li2020}. This enables continuous function evaluation, which makes it particularly well-suited for super-resolution. Indeed, various neural operator architectures have shown promise in spatial \cite{Wei2023,Han2024} and temporal \cite{Zhang2024} super-resolution tasks, but their application to SSR remains limited. While prior work, such as Zhang et al.~\cite{RSNO}, has explored the integration of neural operator architectures for SSR, the direct application of neural operators for SSR remains largely unexplored, with no prior work explicitly formulating the problem as an operator learning problem.

In this paper, SSR is cast as an operator learning problem and addressed using a neural operator framework. The key contributions are as follows.
\begin{itemize}
    \item The SSR problem is formulated as an operator learning problem, laying the mathematical groundwork for applying neural operators to SSR. 
    \item Spatial-Spectral Residual deep Operator Network (SSRON) is proposed, which applies the Deep Operator Network (DeepONet) architecture to SSR while using a spatial–spectral residual CNN as the branch network--adapting architectural principles proven effective in SSR to the operator learning framework. While CNNs have been used in the branch networks of DeepONets \cite{Mei2024, Guo2024}, this paper, to the best of the author’s knowledge, is the first to adapt a spatial–spectral residual CNN architecture to the operator-learning framework for spectral super-resolution.
    \item The strength of SSRON in SSR is demonstrated, achieving improved performance compared to state-of-the-art methods. The application of this architecture to zero-shot SSR for unseen bands during training and its potential to leverage infinite-dimensional mappings is also demonstrated.
\end{itemize}

\section{Problem Formulation}

The sensor calibrated radiance value $L(\lambda)$ can be written as \cite{Du2023}\[
L(\lambda, \vec{x}) = I(\lambda,\vec{x}) R(\lambda, \vec{x}),
\] where $I$ denotes solar radiation and $R$ is the reflectance of the region that the sensor takes its value from. Both the MSI and HSI are downsampled from this continuous function, with 
\begin{align*}
    L_m(i, \vec{x}) = \int L(\lambda, \vec{x})g_{m}(i, \lambda)d\lambda\\
    L_h(i, \vec{x}) = \int L(\lambda, \vec{x})g_{h}(i, \lambda)d\lambda,
\end{align*} where $g_{m}(i, \lambda), g_{h}(i, \lambda)$ are the Spectral Response Function (SRF) of $i$th band of the multispectral and hyperspectral instruments, respectively. Because the hyperspectral instruments' SRF are centered about a central wavelength $\lambda_i$, we can approximate, 
\begin{gather*}
L_h(\lambda_i, \vec{x}) \approx L(\lambda_i, \vec{x})\int g_{h}(i, \lambda)d\lambda =  L(\lambda_i, \vec{x}).
\end{gather*}
Now consider the functional \[L'(\vec{g}, \vec{x}) \coloneq\int L(\lambda, \vec{x})\vec{g}(\lambda)d\lambda= \vec{L_g}(\vec{x}).\]
Effectively, the MSI can be written as $L'(\vec{g_m}, \vec{x})$. The SSR problem, therefore, can be framed as an operator learning problem that learns $G: L'(\vec{g}, \vec{x})\mapsto L(\lambda, \vec{x})$. In this study, we solve a specific case of the problem where $g$ is fixed to some $g_m$ though extension of this problem to multiple $g$s are possible.

\section{Methodologies}
\subsection{Architecture}
\begin{figure*}[!htb]
    \centering
    \begin{tabular}{cc} 
        \includegraphics[width=0.5\linewidth]{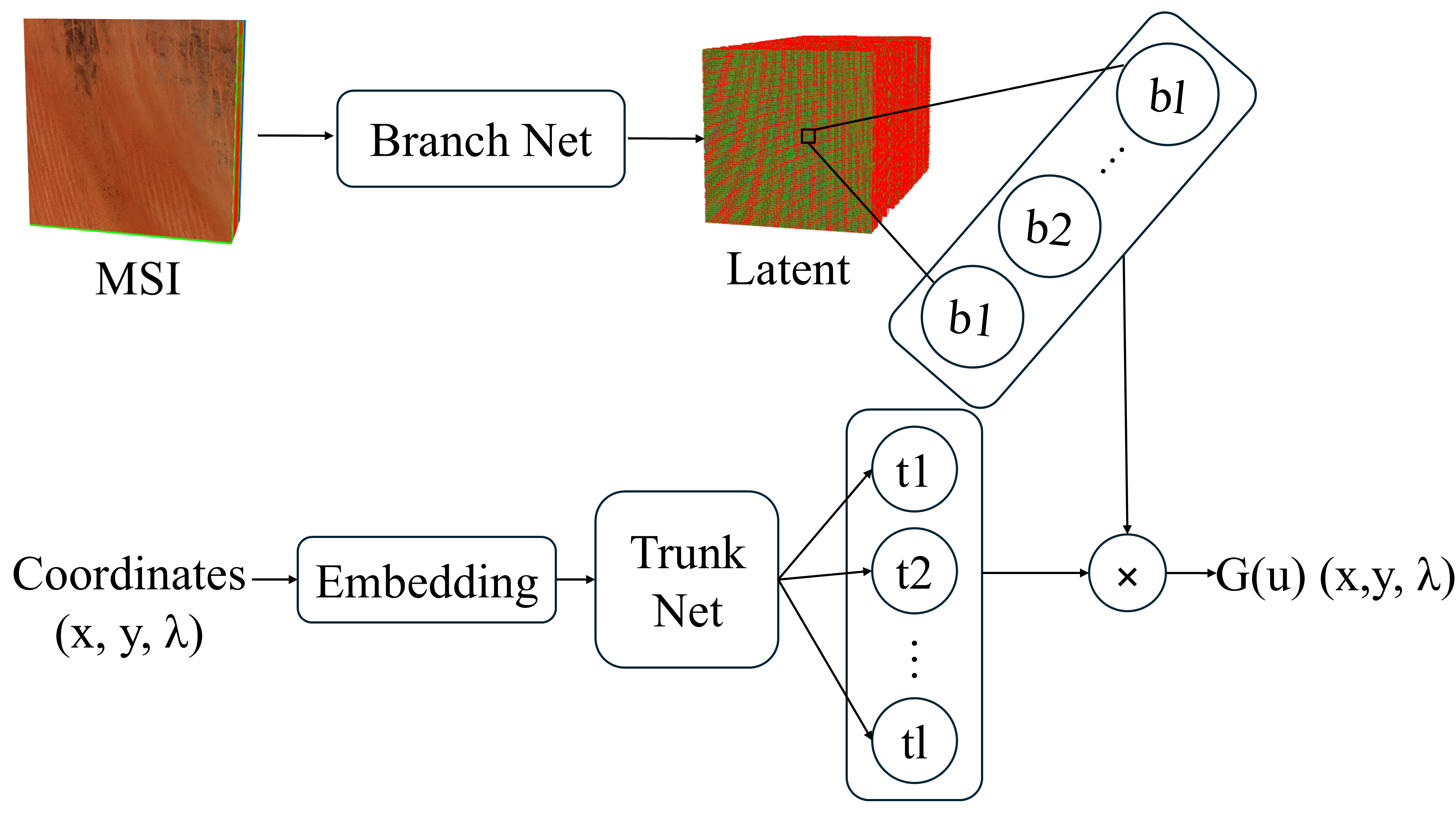}&
         \includegraphics[width=0.5\linewidth]{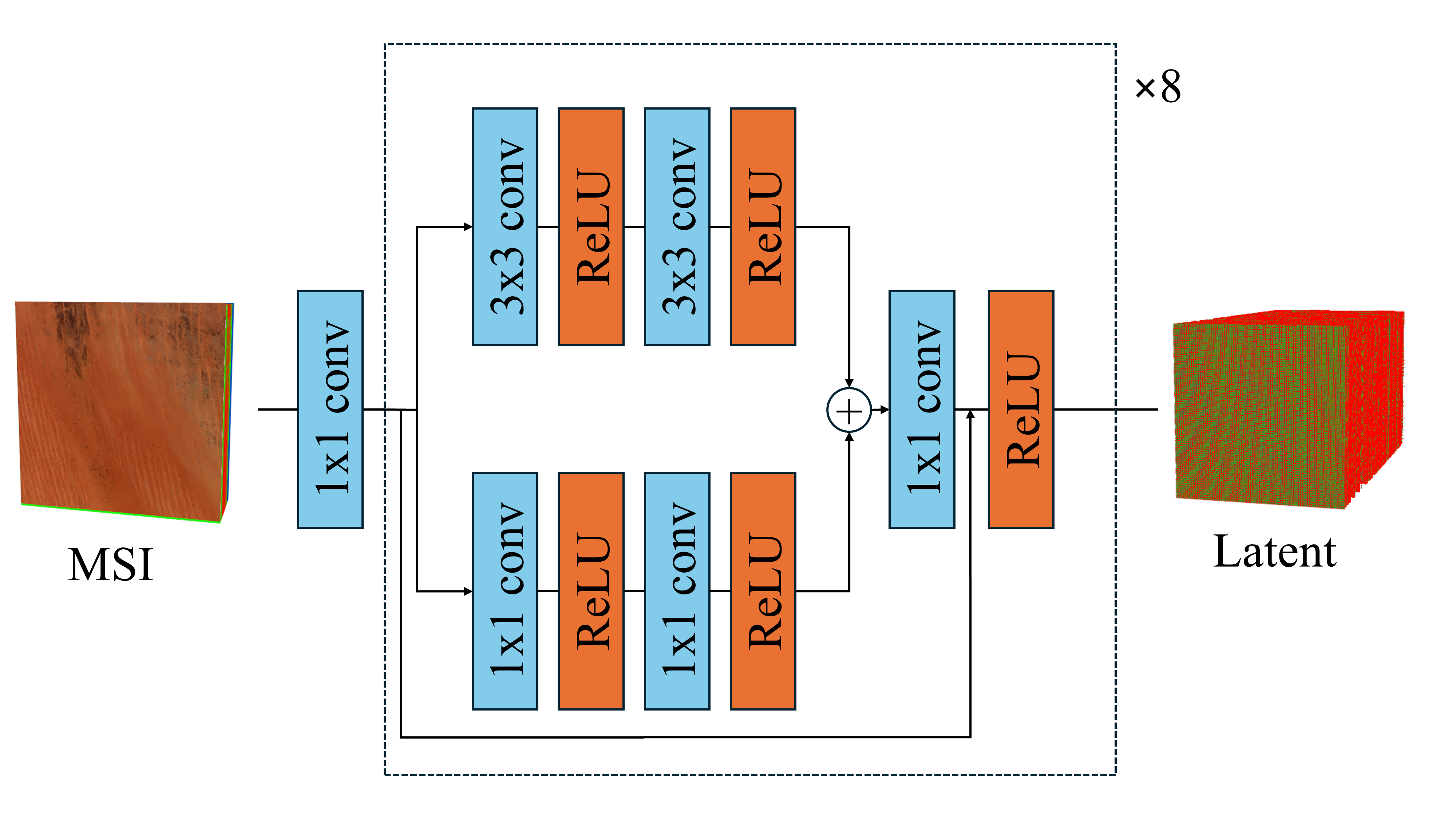}\\
         (a)& (b) \\
    \end{tabular}
    \caption{Proposed architecture of SSRON. The framework consists of a branch network and trunk network, as shown in (a). The branch network encodes the MSI into a latent representation, and the trunk network encodes the spatial-spectral query positions. (b) shows the specific architecture of the branch network.}
    \label{fig:architecture}
\end{figure*}

The proposed architecture is based on the DeepONet \cite{Lu2021}, which creates a branch network that encodes the input function at discrete sensor points and a trunk network that encodes locations where the output is evaluated. The overall architecture is outlined in Figure \ref{fig:architecture}. 

The standard multi-layer perceptron branch network is replaced with a spatial–spectral residual CNN, motivated by the strong performance of such architectures in remote sensing SSR~\cite{SSRAN, Shu2023,Atik2024}.

The embedding layer in Figure \ref{fig:architecture}a maps each coordinate value to a vector according to the following: 
\[
\text{Embedding}: c \mapsto 
\begin{bmatrix}
c \\
\sin(f_0 \cdot c) \\
\cos(f_0 \cdot c) \\
\vdots \\
\sin(f_j \cdot c) \\
\cos(f_j \cdot c)
\end{bmatrix}\] 
where $f_i=2^i$. The coordinates $(x,y,\lambda)$ are each embedded independently, and the resulting vectors are concatenated. The trunk network is a simple two-layer fully-connected neural network with 256 hidden dimensions.  
\subsection{Dataset}
The EMIT Imaging Spectrometer's L1B at sensor calibrated radiance~\cite{EMITdata} was used as the HSI source. EMIT is an advanced imaging spectrometer instrument on the International Space Station that outputs 285 discrete bands from the visible to the short-infrared. Bands 74-79, 100-107, 131-143, and 191-217 were dropped due to absorption via water vapor and ozone, leaving a total of 229 bands. 10 scenes were randomly selected from December 2024 with zero cloud coverage, downloaded from NASA's Earthdata Search. Details of the selected scenes are presented in Table \ref{tab:dataset}. 

\begin{figure}[H]
    \small
    \centering
    \includegraphics[width=\linewidth]{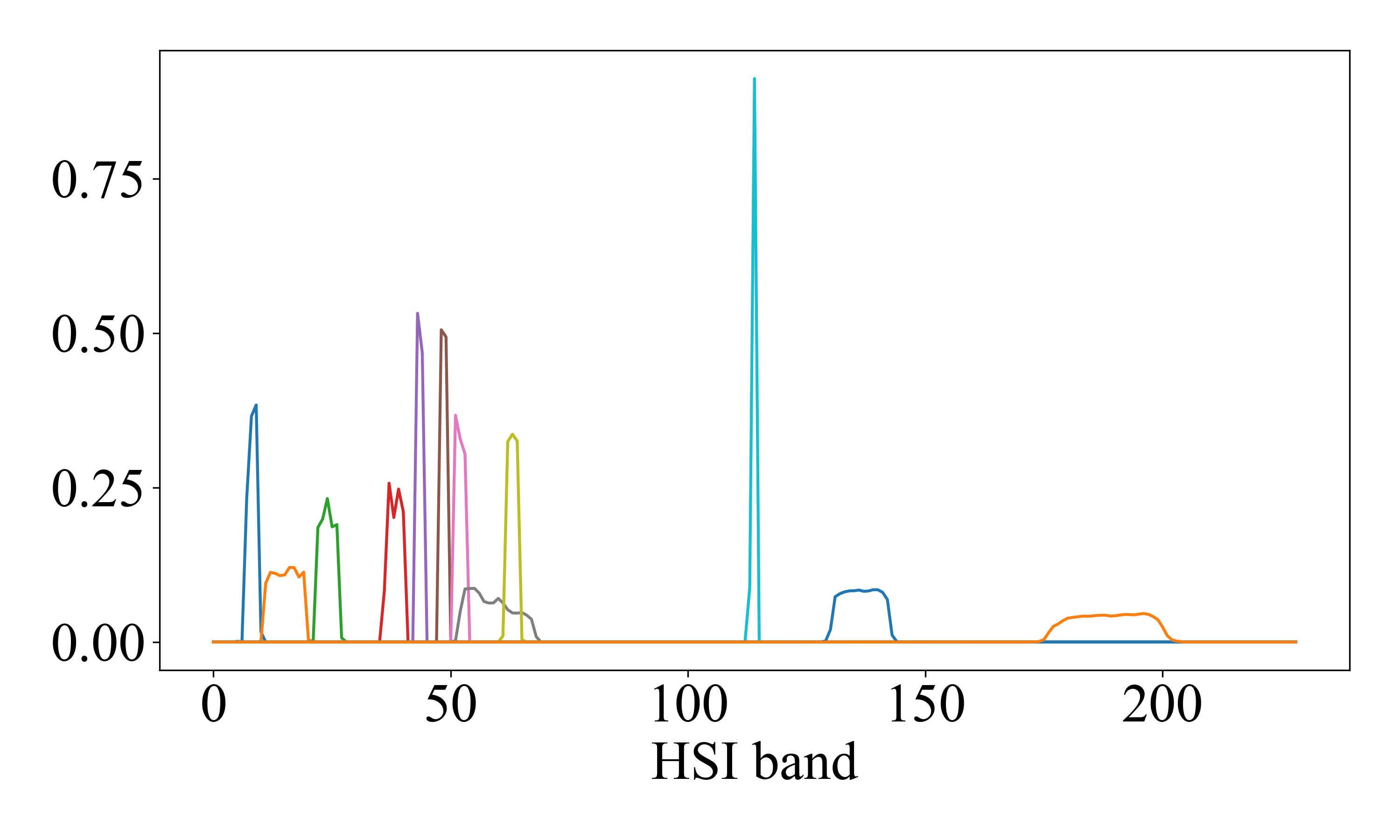}
    \vspace{-15pt}
    \caption{Normalized and downsampled SRF of the Sentinel 2A satellite}
    \label{fig:SRF}
\end{figure}

Spatio-temporally aligned MSIs are generated by downsampling the HSI using the spectral response function of the Sentinel 2A satellite \cite{S2ASRF}. The initial SRF was downsampled to the 229 bands and normalized, following the approach shown in \cite{Li2022}. Band 9 was dropped because it samples corrupted bands in the HSI. The downsampled SRF is shown in Figure \ref{fig:SRF}.

\begin{table}[!htb]
    \centering
    \small
    \caption{Dataset specifications}
    \begin{tabular}{cccccccccc}
        \specialrule{1pt}{2pt}{2pt}
        Scene no.& Granule ID (excluding common prefix) \\\midrule
1&20241214T150837\_2434910\_009\\
2&20241214T151321\_2434910\_033\\
3&20241215T094506\_2435006\_010\\
4&20241215T203042\_2435013\_020\\
5&20241216T071956\_2435105\_007\\
6&20241216T193914\_2435113\_006\\
7&20241217T093419\_2435206\_006\\
8&20241223T050524\_2435803\_051\\
9&20241230T150153\_2436510\_038\\
10&20241231T141323\_2436609\_033\\
        \specialrule{1pt}{2pt}{2pt}
    \end{tabular}
    \label{tab:dataset}
\end{table}
\vspace{-15pt}

\subsection{Training}

Non-overlapping 16x16 patches were selected from each scene, resulting in a total of 61,600 patches. An 8:1:1 train-validation-test split was used. The models were trained with an Adam optimizer at a learning rate of 5e-5, with cosine annealing and early stopping with patience of 5 epochs for 300 epochs, using Mean Absolute Error (MAE) as the loss function. All training was conducted on a single RTX 5060 Ti GPU. 

\subsubsection{Training SSRON} Unlike conventional image-to-image models, the SSRON is trained to predict the value of the learned function at a specified spatial and spectral coordinates. To enable this, 2048 possible positions (spatial and spectral) were selected in the HSI label to be predicted by the SSRON for each epoch. 

\subsubsection{Zero-shot super resolution}
As neural operators learn maps between infinite-dimensional function spaces, they can be evaluated at previously unseen coordinates. To test this capacity, a fixed percentage of bands was intentionally excluded from training, and the model's performance was evaluated across all bands. The excluded bands were selected to be relatively equidistant and to span the same wavelength range as the data. 
\subsection{Baselines}
AWAN, Restormer, and SSRAN architectures were selected as state-of-the-art baselines because they have demonstrated strong performance for SSR tasks~\cite{SSRAN,Restormer,Awan}. The U-shaped Neural Operator (UNO) \cite{rahman2023uno} and the Fourier Neural Operator (FNO) \cite{Li2020} were included as comparable baselines for operator learning. The RSNO proposed by Zhang et al. \cite{RSNO} was not included as a baseline because it relies on additional physics-based information as input to the model, which was neither available nor assumed in this experimental setting.
\subsection{Metrics}
In addition to MAE, Root Mean Squared Error (RMSE), Peak Signal to Noise Ratio (PSNR), and Structural Similarity Score (SSIM) were used as evaluation metrics. SSIM is calculated like so \cite{ZhouWang2004}:
\[
\text{SSIM}(x, y) = \frac{(2\mu_x \mu_y + \epsilon_x)(2\sigma_{xy} + \epsilon_y)}{(\mu_x^2 + \mu_y^2 + \epsilon_x)(\sigma_x^2 + \sigma_y^2 + \epsilon_y)},
\]

where \(\mu_x\) and \(\mu_y\) are the mean values of images \(x\) and \(y\), \(\sigma_x^2\) and \(\sigma_y^2\) are the variances, $\epsilon_x=(0.01\cdot L)^2,\epsilon_y =(0.03\cdot L)^2$ are small constants to stabilize division with L being the dynamic range of data, and \(\sigma_{xy}\) is the covariance of \(x\) and \(y\). 
\FloatBarrier
\section{Results}

Table \ref{tab:main results} reports the results evaluated over the testing dataset. SSRON outperforms all baseline models across all tested metrics. UNO also performs strongly, outperforming all non-operator baseline models. In contrast, the FNO shows substantially higher errors, which may be attributed to the sharpness in spectra that needs to be reconstructed; standard FNOs struggle with such high-dimensional features \cite{Liu2024}. 
\begin{table}[!b]
    \centering
    \small
    \caption{Error metrics over testing dataset}
    \begin{tabular}{cccccccccc}
        \specialrule{1pt}{2pt}{2pt}
        Model type& MAE $\downarrow$ & RMSE $\downarrow$ & PSNR $\uparrow$ & SSIM $\uparrow$\\\midrule
        AWAN & 0.01529 & 0.03476 & 50.92 & 0.9861
          \\
        FNO & 0.1690& 0.2745& 31.23& 0.2516\\
        Restormer & 0.01606& 0.03670& 50.24& 0.9845\\
        SSRAN & 0.01871 & 0.03999 & 49.29 & 0.9820\\
        UNO & 0.01498 & 0.03255 & 51.01 & 0.9865\\
        SSRON & \textbf{0.01183}& \textbf{0.02599}& \textbf{52.67}& \textbf{0.9889}\\
        \specialrule{1pt}{2pt}{2pt}
    \end{tabular}
    \label{tab:main results}
\end{table}

\begin{figure}[H]
    \centering
    \includegraphics[width=\columnwidth]{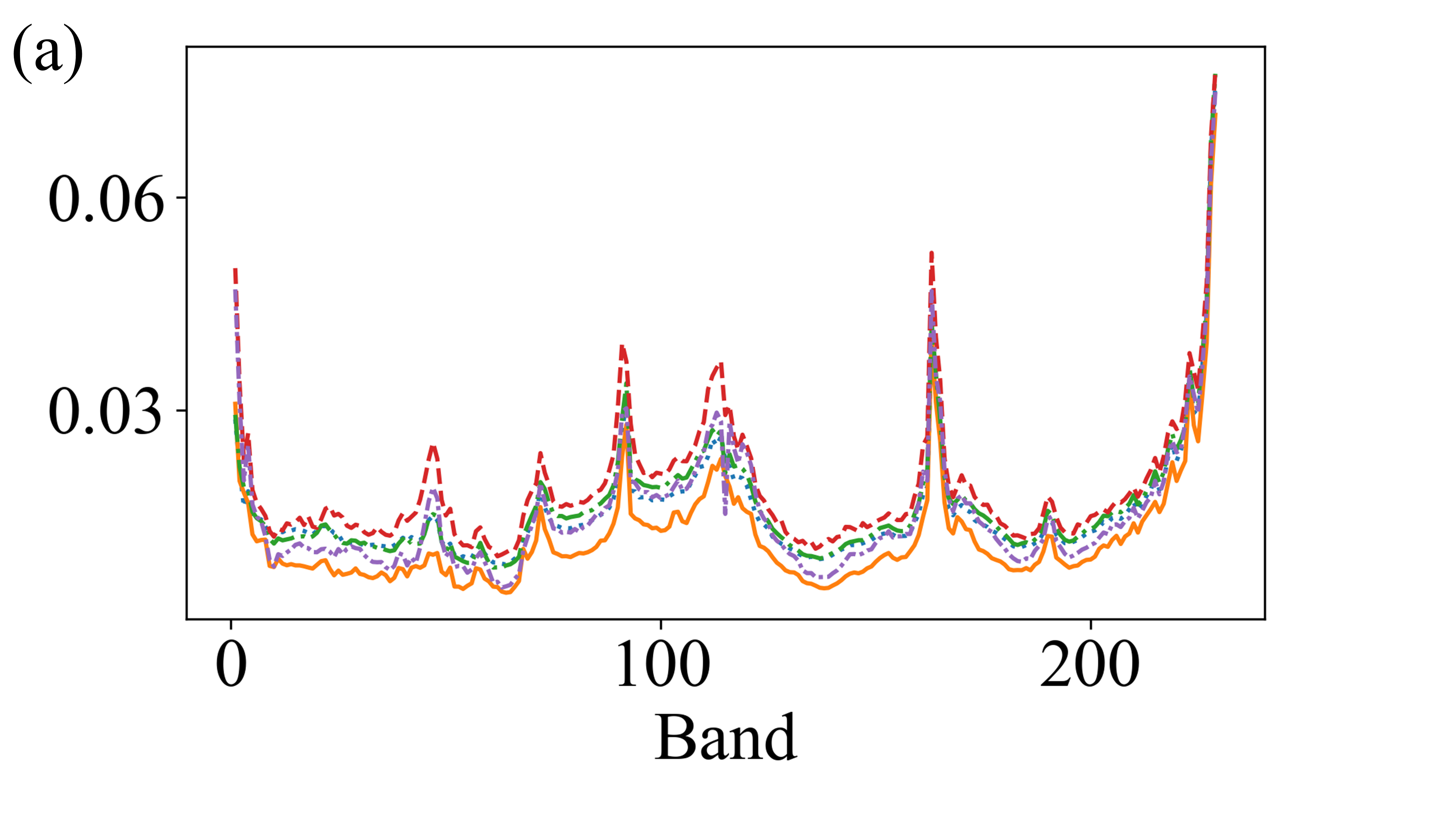}
    \includegraphics[width=\columnwidth]{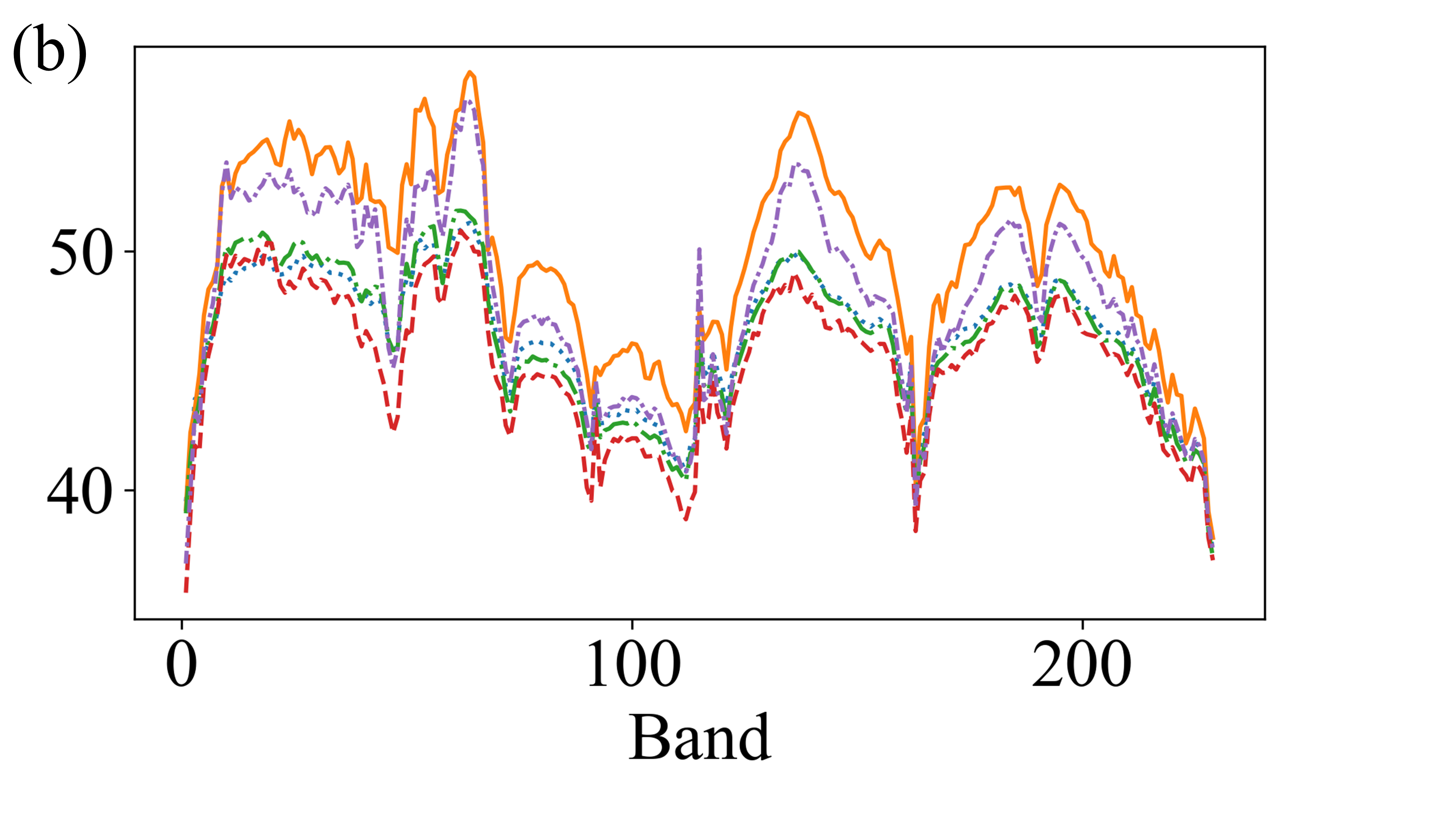}
    \includegraphics[width=\columnwidth]{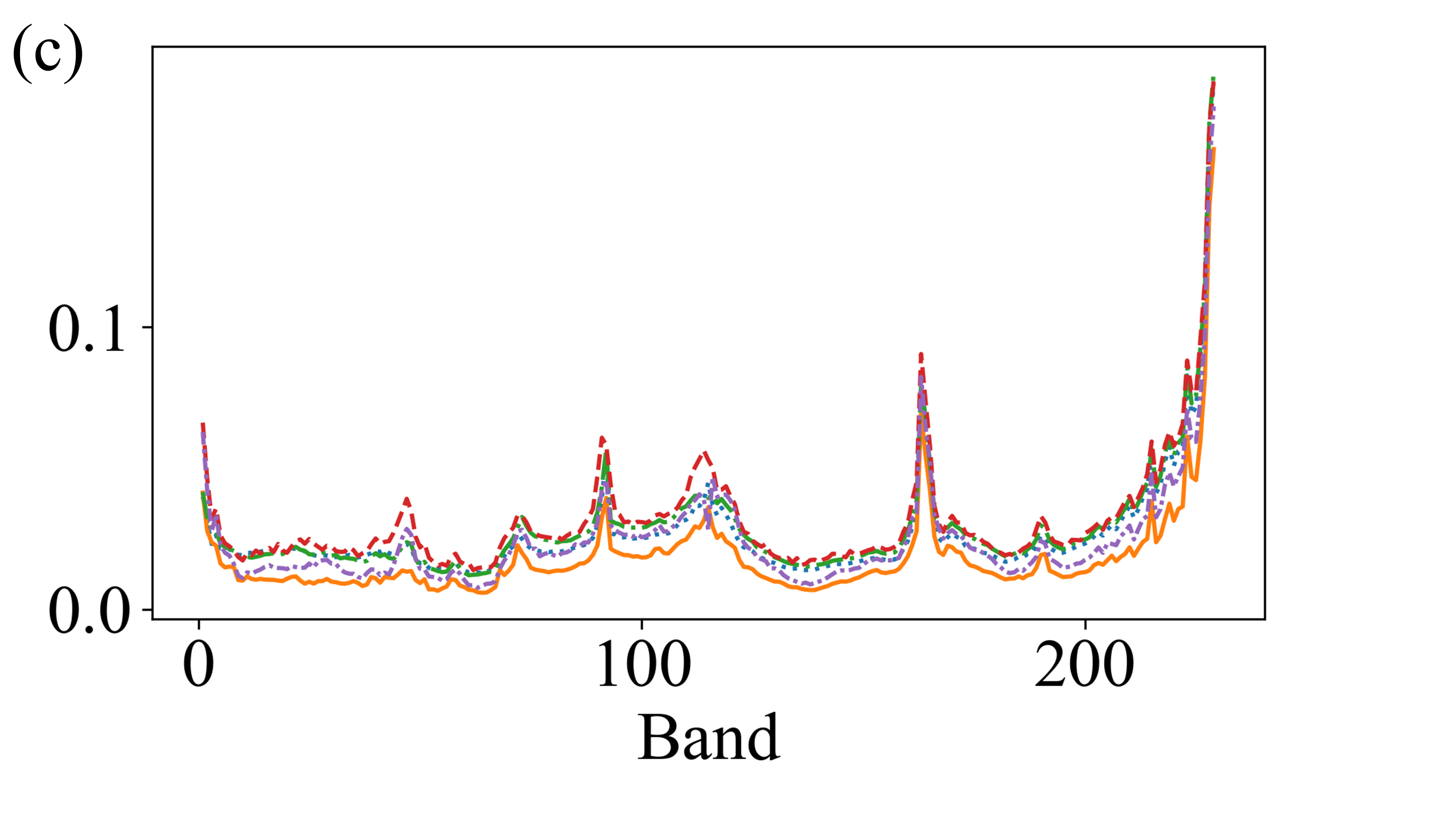}
    \includegraphics[width=\columnwidth]{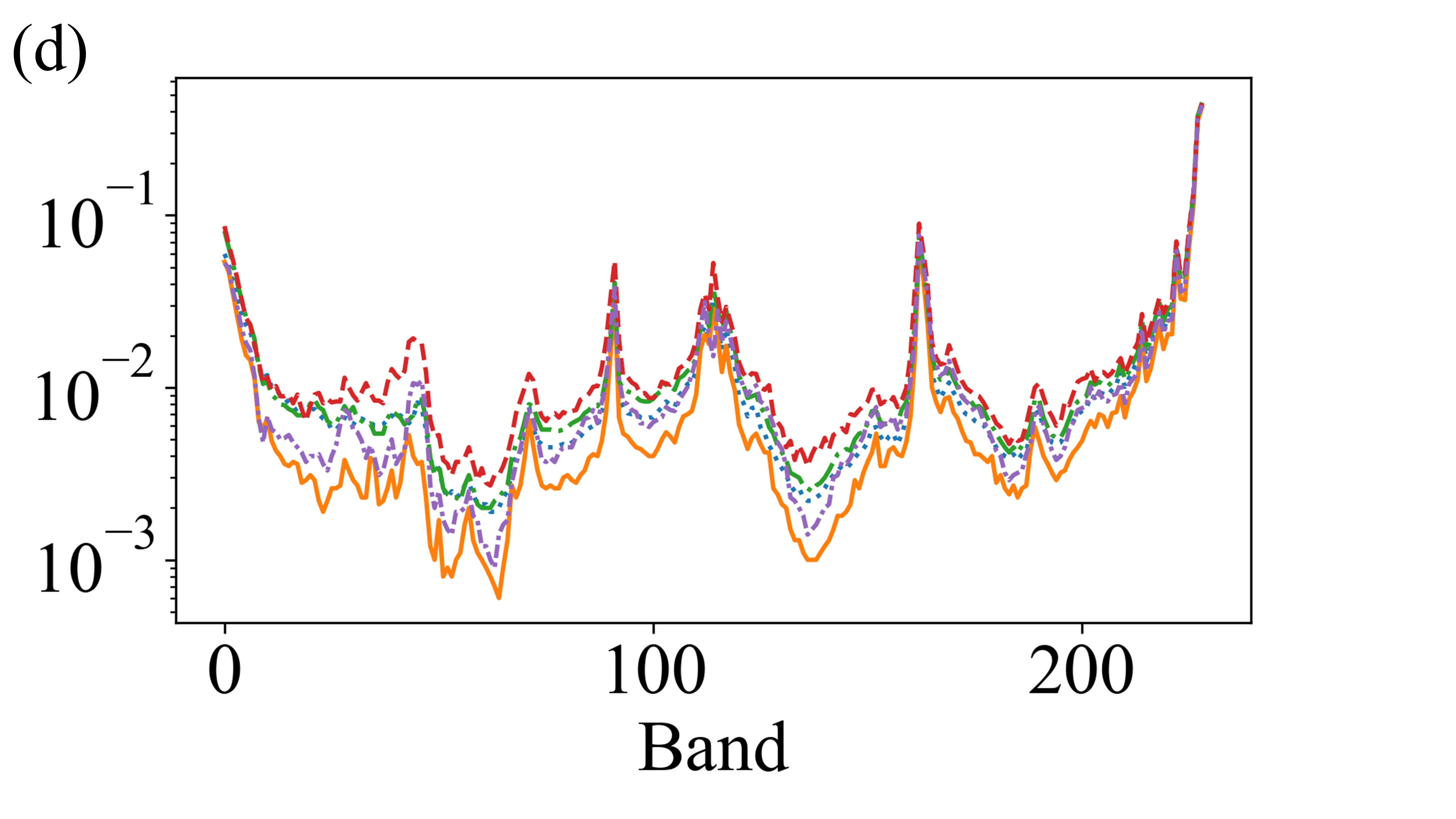}
    \includegraphics[width=\columnwidth]{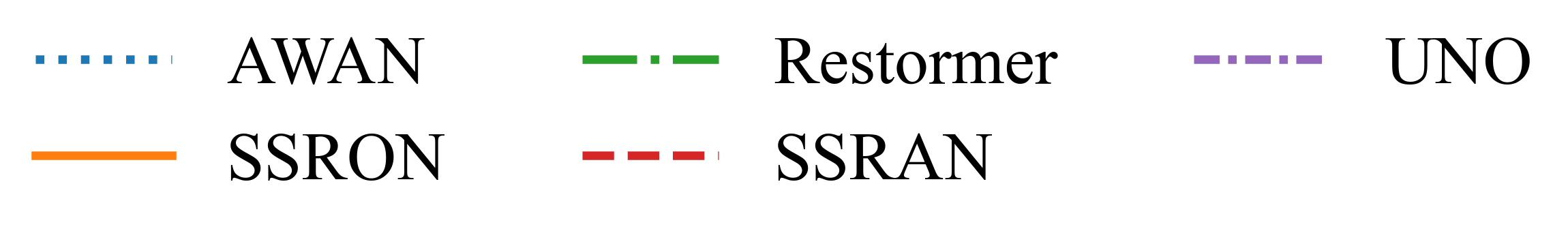}
    \caption{Per-band (a) MAE, (b) PSNR, (c) RMSE, and (d) \newline 1-SSIM value comparison. FNO is excluded from the analysis due to extremely high errors. SSIM is reported as 1-SSIM and on a log scale for enhanced visibility. }
    \label{fig:pba}
\end{figure}

Figure \ref{fig:pba} shows the per-band errors of each model. SSRON achieves a lower error across all bands, with the exception of band 115, where the UNO shows a lower error. A sharp increase in error is observed in the bands below 5 and above 200. This behavior is likely due to the limited spectral information of the Sentinel-2A satellite in those regions, as shown in Figure \ref{fig:SRF}. 

The relationship between SRF coverage and the errors is further supported by Pearson's correlation test. The correlation between the MAE ($\downarrow$), RMSE($\downarrow$), and PSNR($\uparrow$) with the SRF value at each wavelength is -0.222, -0.175, and 0.408 (p $<$ 0.01), respectively, demonstrating a clear inverse relationship between error and SRF values. 

Table \ref{tab:zero-shot} reports the zero-shot spectral resolution performance of the SSRON. As the percentage of bands used decreases, MAE increases and SSIM decreases, indicating that zero-shot inference is difficult. RMSE and PSNR do not strictly follow the monotonic trend, with the 75\% model
demonstrating a lower RMSE than the 80\% model and the 60\% model demonstrating a lower RMSE and higher PSNR than the 65\% model. Such behavior likely relates to which bands were removed for training. If bands with higher errors in figure \ref{fig:pba} were removed, the model would not be penalized for poor performance for those bands, increasing the error of the final trained model.

\begin{table}[!tb]
    \centering
    \small
    \caption{Error metrics over testing dataset with reduced number of bands seen during training}
    \begin{tabular}{cccccccccc}
        \specialrule{1pt}{2pt}{2pt}
        \% bands used& MAE $\downarrow$ & RMSE $\downarrow$ & PSNR $\uparrow$ & SSIM $\uparrow$\\\midrule
        100 & 0.01183 & 0.02599 & 52.67& 0.9889 \\
        95 & 0.01422 & 0.04756 & 48.02 & 0.9764 \\
        90 & 0.01509 & 0.04836 & 47.50 & 0.9748 \\
        85 & 0.01701 & 0.05290 & 46.64 & 0.9730 \\
        80 & 0.01844 & 0.05638 & 46.10 & 0.9716 \\
        75 & 0.01946 & 0.05599 & 45.64 & 0.9707 \\
        70 & 0.02077 & 0.07779 & 43.27 & 0.9690 \\
        65 & 0.02615 & 0.08810 & 41.92 & 0.9644 \\
        60 & 0.02837 & 0.08463 & 42.32 & 0.9620 \\
        55 & 0.03239 & 0.1046 & 40.62 & 0.9557 \\
        50 & 0.03929 & 0.1075 & 39.84 & 0.9502 \\
        \specialrule{1pt}{2pt}{2pt}
    \end{tabular}
    \label{tab:zero-shot}
\end{table}
Nevertheless, SSRON remains competitive with the state-of-the-art performance at 90\% of bands used and attains a lower MAE than other methods. Because SSRON takes wavelengths as a continuous input, the model can in-principle be evaluated between the native sensor bands. The zero-shot results suggest the potential for such spectral interpolation, although validation against denser spectral measurements would be needed to confirm the accuracy of HSI beyond the native discretization.

\section{Conclusion and future directions}

This work presents SSRON, a deep operator network that uses spatial-spectral convolutions with residual connections as its branch network. Empirically, the SSRON achieves state-of-the-art accuracy on simple SSR and demonstrates its potential for zero-shot prediction of additional bands.

The paper offers a novel perspective on the SSR task and opens a new direction for applications of operator learning. Future work should focus on optimizing neural operator architectures to further improve model performance. In addition, the continuous spectral representation enabled by this framework motivates further study of the densely sampled spectral interpolation for downstream applications, such as materials classification of imaged regions. Finally, varying the input SRF to support the general application of SSRON to various satellites is an important direction for future work.
\section{Acknowledgements}
The author is grateful to receive funding from the George W. Housner Student Discovery Fund for support in the research and attendance at the conference. The author would like to thank Enes Banushi for productive discussion on the problem formulation. 
\small
\newpage
\bibliographystyle{IEEEtranN}
\bibliography{references}

\end{document}